\documentclass[letterpaper, 10 pt, conference]{ieeeconf}  

\IEEEoverridecommandlockouts   

\usepackage{graphics} 
\usepackage{epsfig} 
\usepackage{times} 
\usepackage{multicol}
\usepackage[bookmarks=true]{hyperref}
\usepackage{array}
\usepackage{hyperref}
\usepackage{amsmath}
\usepackage{cleveref}
\usepackage{paralist}
\usepackage{float}
\usepackage{pdfpages}
\let\labelindent\relax
\usepackage{enumitem}
\usepackage{wrapfig}
\usepackage{booktabs} 
\usepackage{multirow}
\usepackage{hyperref}
\usepackage{url}
\usepackage{amssymb}
\usepackage{graphicx}
\usepackage{CJKutf8}
\usepackage{algorithm}
\usepackage[noend]{algorithmic}
\usepackage{colortbl}
\usepackage{threeparttable}
\usepackage{mdframed}
\usepackage{tcolorbox}
\tcbuselibrary{minted}
\usepackage{blindtext}
\usepackage{multicol}
\usepackage{caption, subcaption}
\usepackage[utf8]{inputenc}
\usepackage{tcolorbox}
\usepackage{listings}
\usepackage{xcolor}
\usepackage{tabularx}
\usepackage{cite}

\definecolor{customblue}{HTML}{ccf2f5}
\usepackage{soul, color, xcolor}

\newcommand{\ours}[0]{AdaptPNP}

\title{\LARGE \bf AdaptPNP: Integrating Prehensile and Non-Prehensile Skills \\for Adaptive Robotic Manipulation}

\author{Jinxuan Zhu$^{*}$$^{1,2}$, Chenrui Tie$^{*}$$^{1}$, Xinyi Cao$^{*}$$^{2,3}$, Yuran Wang$^{4}$, Jingxiang Guo$^{1}$, \\Zixuan Chen$^{1,5}$, Haonan Chen$^{1}$, Junting Chen$^{1}$, Yangyu Xiao$^{2}$, Ruihai Wu$^{4}$ and Lin Shao$^{\dagger}$$^{1, 2}$
\thanks{*~Equal contribution.}%
\thanks{$\dagger$~Corresponding author: Lin Shao (\texttt{linshao@nus.edu.sg}).}%
\thanks{$^{1}$ School of Computing, National University of Singapore, Singapore.}%
\thanks{$^{2}$ RoboScience, Beijing, China.}%
\thanks{$^{3}$ East China Normal University, Shanghai, China.}%
\thanks{$^{4}$ Peking University, Beijing, China.}%
\thanks{$^{5}$ Nanjing University, Nanjing, China.}%
}

\begin{document}

\maketitle

\thispagestyle{empty}
\pagestyle{empty}


\begin{abstract}
Non-prehensile (NP) manipulation, in which robots alter object states without forming stable grasps (for example, pushing, poking, or sliding), significantly broadens robotic manipulation capabilities when grasping is infeasible or insufficient. 
However, enabling a unified framework that generalizes across different tasks, objects, and environments while seamlessly integrating non-prehensile and prehensile (P) actions remains challenging: robots must determine when to invoke NP skills, select the appropriate primitive for each context, and compose P and NP strategies into robust, multi-step plans.
We introduce \ours, a vision-language model (VLM)-empowered task and motion planning framework that systematically selects and combines P and NP skills to accomplish diverse manipulation objectives. 
Our approach leverages a VLM to interpret visual scene observations and textual task descriptions, generating a high-level plan skeleton that prescribes the sequence and coordination of P and NP actions. 
A digital-twin based object-centric intermediate layer predicts desired object poses, enabling proactive mental rehearsal of manipulation sequences. 
Finally, a control module synthesizes low-level robot commands, with continuous execution feedback enabling online task plan refinement and adaptive replanning through the VLM.
We evaluate \ours\ across representative P\&NP hybrid manipulation tasks in both simulation and real-world environments. 
These results underscore the potential of hybrid P\&NP manipulation as a crucial step toward general-purpose, human-level robotic manipulation capabilities.
More detailed information can be found at \href{https://adaptpnp.github.io/}{https://adaptpnp.github.io/}.





\end{abstract}

\section{Introduction}
When manipulating objects to achieve desired configurations, robots typically rely on establishing stable grasps and transporting objects to target locations. However, prehensile manipulation often becomes infeasible or insufficient due to potential collisions, or object characteristics that preclude stable grasping. 
For instance, when transferring a thin card lying flat on a tabletop that cannot be grasped directly, a robot could first push it toward the table edge, then grasp and lift it from the side. Such scenarios necessitate non-prehensile (NP) manipulation strategies\cite{ruggiero2018nonprehensile}, where robots alter object states through contact interactions like pushing, poking, or sliding without forming stable grasps. The integration of prehensile and non-prehensile skills significantly expands robotic manipulation capabilities, enabling more versatile and human-like object interactions.

However, effectively incorporating non-prehensile skills into robotic manipulation frameworks presents significant challenges.
First, the diversity of NP primitives ($e.g.$, pushing, poking, pressing, sliding) dramatically increases manipulation dexterity but also creates combinatorial complexity in skill selection and sequencing. 
Second, robots must reason about and compose P\&NP actions based on environmental constraints, object properties, and task requirements, which demands sophisticated understanding of skill preconditions, effects, and inter-dependencies.
Third, this combinatorial explosion is compounded by the inherently multi-modal nature of manipulation tasks: numerous skill combinations may appear semantically reasonable, yet only a subset proves physically feasible or execution-efficient under physical constraints.
Together, these factors complicate the generation of robust, executable task plans that balance semantic plausibility, physical feasibility, and execution reliability.

\begin{figure}[!t]
\centering
\includegraphics[width=0.9\linewidth]{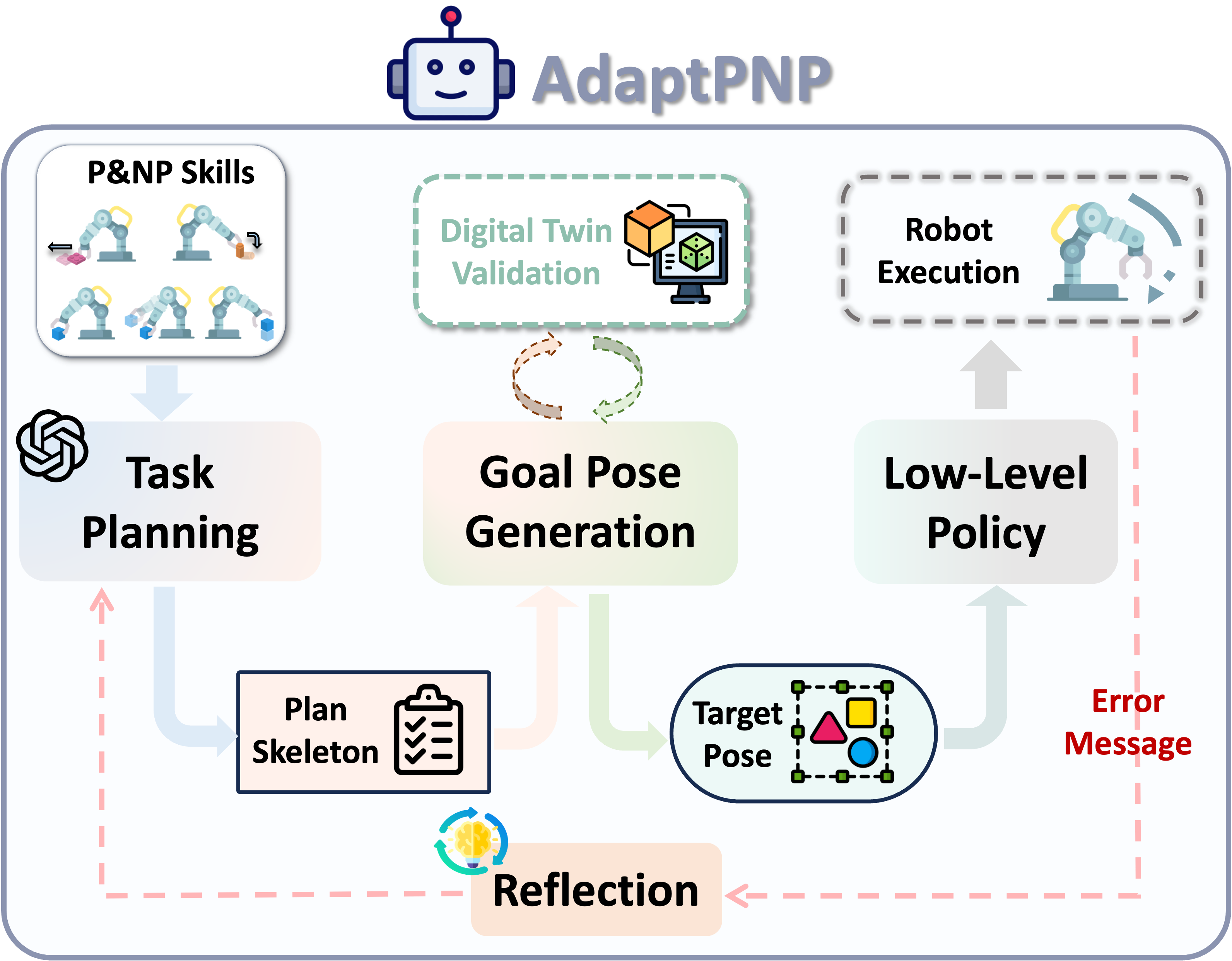}
\caption{\textbf{Overview of AdaptPNP.} A VLM-based task planner generates a mixed sequence of prehensile (grasp, moveto, release) and non-prehensile (push, rotate) primitives; a digital twin ``mentally rehearses'' each primitive by generating target 6D object poses; and a closed-loop reflection mechanism uses execution feedback to iteratively refine the plan.}
\vspace{-3mm}
\label{fig:teaser}
\end{figure}

Existing methods often lack adaptive support for integrating prehensile and non-prehensile skills.
Single-task NP policies~\cite{cho2024corn,wang2025dexterous, ding2024preafford,lee2024non,lyu2025dywa} excel at isolated primitives like pushing or sliding, but cannot choose among multiple NP options or integrate grasping based on scene context. 
Classical task and motion planning (TAMP) frameworks~\cite{aeronautiques1998pddl, garrett2020pddlstream} sequence diverse actions using symbolic models ($e.g.$, PDDL~\cite{aeronautiques1998pddl}) rely on extensive hand-crafted domain knowledge, limiting adaptability to novel objects and environments. 
VLM-based planners represent tasks as natural-language subgoals~\cite{liu2024moka}, image-space waypoints~\cite{yuan2024robopoint} or spatial relations between 3D points~\cite{huang2024rekep}, which cannot fully specify NP action in all spatial directions ($e.g.$, poking to induce rotation) and often omit closed-loop plan refinement.
A general P\&NP framework must adaptively integrate P\&NP skill according to visual and physical features of task, object and environment.


To address these challenges, we present \ours, an integrated task-and-motion planning framework that unifies high-level P\&NP skill scheduling with low-level execution. 
At its core, a VLM takes an RGB scene and natural-language task instruction to infer a sequence of prehensile and non-prehensile primitives,  leveraging commonsense knowledge for zero-shot generalization to novel objects and environments. For each primitive, multiple candidate 6DoF subtarget object poses are generated according to task instruction and filtered through a digital twin to discard physically infeasible options, providing precise intermediate objectives for NP actions. 
Selected subtargets feed into off-the-shelf motion planning module to execute the corresponding motions. 
Crucially, after executing each skill, visual observations and controller feedback provide physical insights to the VLM task planner, enabling it to identify physically feasible skill sequences among multiple semantically plausible alternatives and resolve multi-modal ambiguities in P\&NP planning. 

We evaluate \ours\ on eight challenging P\&NP hybrid manipulation tasks in both simulation and real-world settings. 
\ours\ consistently outperforms non-hierarchical reinforcement-learning and MPC approaches, hierarchical VLM-based planning methods, and end-to-end vision-language-action models. 
These results demonstrate that \ours\ exactly expands robotic capabilities and effectively handles diverse, complex manipulation challenges.

To summarize, our key contributions are:
\begin{itemize}

\item We propose \ours, a generalizable manipulation framework that adaptively selects and coordinates prehensile and non-prehensile skills based on task requirements, object properties, and environmental constraints, enabling flexible skill composition across diverse manipulation scenarios through VLM-guided planning.

\item We introduce a digital-twin based intermediate representation that generates physically-informed 6D object target poses of each skill, bridging the planning-execution gap by capturing both translational and rotational object states while respecting physical constraints.


\item We demonstrate comprehensive evaluation across diverse manipulation scenarios requiring coordinated P\&NP skills, including object alignment, extrinsic dexterity exploitation, and tool using, achieving superior performance over existing approaches.

\end{itemize}

\section{related work}
\subsection{Non-Prehensile Manipulation Skills}
Non-prehensile manipulation tasks can be broadly categorized into three types.
First, contact-based non-grasping manipulation directly changes object states via interactions like pushing~\cite{ding2024preafford,zhong2025activepusher}, poking~\cite{zhou2023hacman, jiang2024hacman++, cho2024corn}, or sliding~\cite{cheng2022contact}, in which robots apply contact forces and friction without ever forming a stable grasp to achieve the desired motion;
Second, extrinsic dexterity leverages environmental features, such as walls, edges, slots, or slopes, to enable complex manipulations~\cite{wang2025dexterous,kim2023pre};
Third, tool-mediated manipulation employs auxiliary tools like hooks~\cite{lee2024non,shao2020learning}, push bars~\cite{gao2025vlmgineer}, or rods and hammers~\cite{jiang2023contact, huang2024copa} to extend the robot's effective reach and manipulation capabilities.
While these approaches demonstrate the effectiveness of individual NP skills, they typically rely on task-specific action pattern and policy, limiting their adaptability to novel tasks and scenarios.

\subsection{Task and Motion Planning}
Task and motion planning (TAMP) seeks to jointly reason over high‐level symbolic actions and low‐level continuous motions to generate executable plans for robotic systems. Early work introduced integrated planners that interleave symbolic search with geometric reasoning, using finite state models and sampling to bridge discrete and continuous domains~\cite{srivastava2014combined,ahmetoglu2025symbolic}. 
More recent approaches formalize TAMP in extensions of PDDL, such as PDDLStream~\cite{garrett2020pddlstream}, which embeds black‐box samplers as ``streams'' to lazily generate continuous parameters during task search. Optimization‐based methods further cast TAMP as a joint discrete‐continuous optimization problem, employing nonlinear solvers to refine action sequences and trajectories simultaneously~\cite{zhao2024survey}. 
Despite these advances, conventional TAMP frameworks often rely on hand‐crafted domain models or struggle with the combinatorial growth of sampling calls, motivating our VLM‐guided, feedback‐driven alternative.

\subsection{Vision-Language Models for Robotic Manipulation}

Recent breakthroughs in vision-language models (VLMs) have sparked interest in open-world and zero-shot robotic manipulation due to their strong visual understanding and commonsense reasoning. Many works leverage VLMs to decompose high-level instructions into subtasks~\cite{ zhao2025learning, tie2025manual2skill} or generate continuous parameters like 2D keypoints, reward functions or action vectors~\cite{nasiriany2024pivot, huang2023voxposer}. 
However, trained primarily on visual question-answering data, VLMs often fail to understand physical interactions, such as object dynamics, collisions, and contact physics, leaving a gap to motion planning. To address this, some methods add environmental feedback or chain-of-thought prompts for online self-correction~\cite{duan2024aha, liu2023reflect}, but they typically decouple planning from execution or ignore low-level errors. 
In contrast, our framework employs a digital twin to generate fully 3D object poses as an intermediate representation, then validates subtargets and feeds execution feedback to the VLM for iterative plan adjustment, enabling seamless integration of semantic reasoning and physical consequence.
\section{Problem Formulation}

Assume we have a set of P\&NP action primitives $\mathbf{S} = \{s_1, ..., s_n\}$, where each primitive carries semantic meaning. While previous work defines diverse NP skills according to their semantic properties ($e.g.$, push, press, slide), we adopt a spatial effect based classification that captures the fundamental geometric transformations achievable through non-prehensile manipulation. For tabletop scenarios, we identify two core non-prehensile primitives that collectively encompass the majority of spatial manipulations:

\begin{itemize}
\item \textit{Push}: Transforms objects within SE(2) space through planar translation and in-plane rotation, enabling repositioning and reorientation on the supporting surface. 
For example, in Figure~\ref{fig:task}(Box), the robot can push to align the box with the visual target.
\item \textit{Rotate}: Induces out-of-plane rotation about contact points or edges, allowing objects to tilt or flip relative to the table surface. For instance, in Figure~\ref{fig:task}(Book), the robot rotates the blue book to retrieve it from the shelf.
\end{itemize}
Furthermore, the framework can naturally expand to more diverse primitive skills.

Also, we define three prehensile primitives:
\begin{itemize}
\item \textit{Grasp}: Establishes stable object-robot connection via parallel gripper closure.
\item \textit{Moveto}: Transports grasped objects to targets.
\item \textit{Release}: Terminates grasp through gripper opening.
\end{itemize}


Inspired by previous works~\cite{ning2025prompting}, we assume access to a digital twin of the execution environment that includes the objects and surrounding terrain, without the robot arm itself.
Such 3D reconstructions, built from multi-view or depth-based sensing, are now both mature and widely available~\cite{wang2025vggt}, providing a reliable virtual workspace in which to sample and validate candidate 6D object poses before execution.


Given a visual observation $\mathcal{O}$ and a natural-language task instruction $\mathcal{I}$ ($e.g.$, ``move \{\textit{Object}\} to \{\textit{Target}\}'') as well as the action primitives $\mathbf{S} = \{s_{1}, \ldots, s_{n}\}$, our method comprises three stages.
First, given the scene observation $\mathcal{O}$ and task instruction $\mathcal{I}$, the VLM-based task planner generates a plan skeleton $\mathcal{P} = [s_1(\lambda_1), \dots, s_m(\lambda_m)]$, a sequence of action primitives $s_i$ where each is paired with discrete parameters $\lambda_i$ that specify the target object and relevant region or pose.
Second, for each primitive $s_i(\lambda_i)$ in $\mathcal{P}$, we generate the corresponding subtarget object pose. Since objects are rigid bodies, each pose is represented by a transformation $T\in SE(3)$.
Third, the motion module takes an instantiated primitive $s_i(\lambda_i)$ along with initial object poses $\{T_{1}^{\text{init}},\dots,T_{k}^{\text{init}}\}$ and target poses $\{T_{1}^{\text{tgt}},\dots,T_{k}^{\text{tgt}}\}$ to synthesize a robot trajectory $\tau$ that executes $s_i(\lambda_i)$ and drives the objects from their initial to target configurations.
\section{method}

\begin{figure*}[!htb]
\centering
\includegraphics[width=1.0\linewidth]{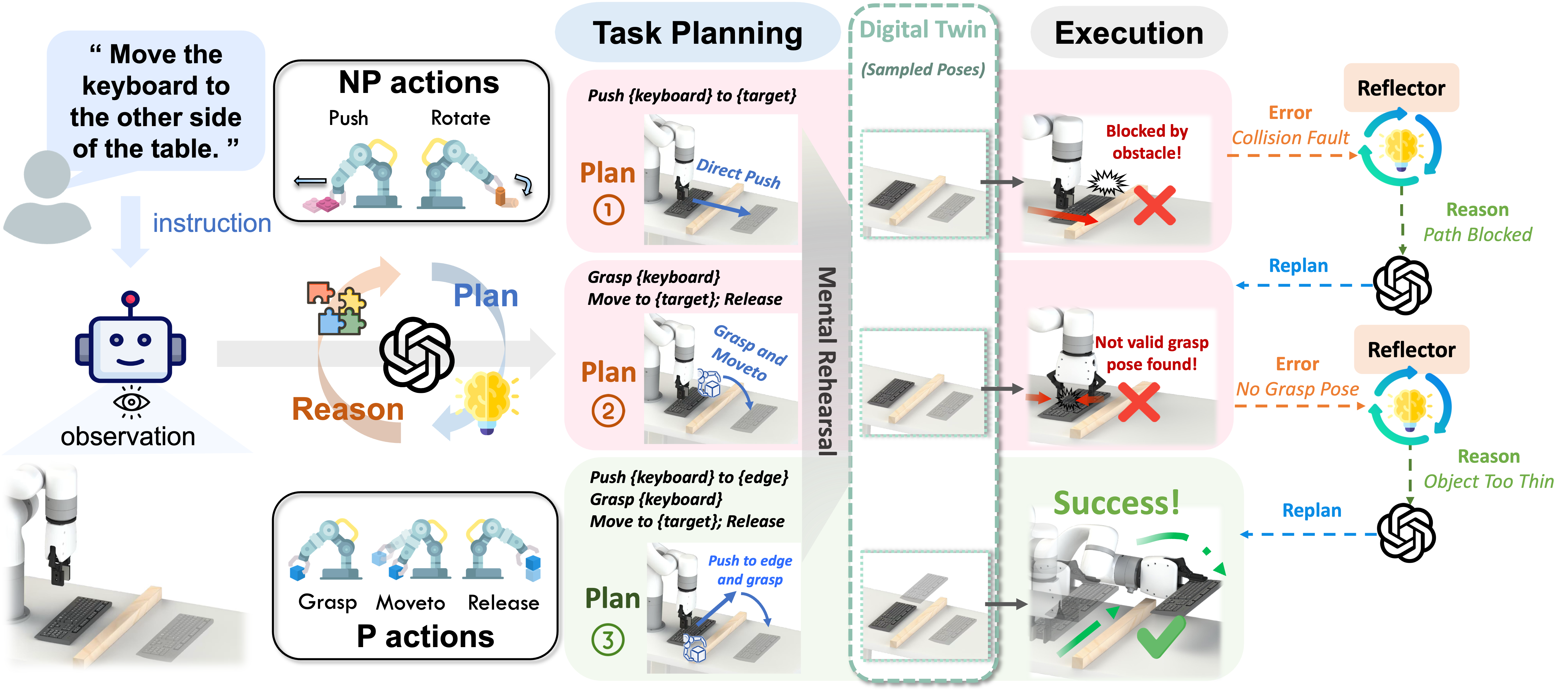}
\caption{\textbf{Pipeline of \ours.} Starting from an instruction and scene image, the task planner generates an initial plan ($e.g.$, direct push), which is mentally rehearsed in the digital twin to sample a 6D target pose. After execution fails, the reflector analyzes the error and provides insight to the planner, which replans ($e.g.$, grasp-and-move). This loop continues until the successful plan (push-to-edge-then-grasp) completes the task.}
\label{fig:pipeline}
\end{figure*}




We propose a unified, three-stage framework for adaptive task-and-motion planning in hybrid prehensile and non-prehensile manipulation. In~\Cref{Sec: Plan Skeleton Generation and Reflection}, we present Plan Skeleton Generation and Reflection, where a VLM-derived sequence of action primitives is instantiated and iteratively refined using execution feedback.
\Cref{Sec: Sub-goal Pose Generation} details Sub-goal Pose Generation, demonstrating how 6D object poses serve as a general intermediate representation that bridges high-level semantic plans and low-level motion objectives. Finally, \Cref{Sec: Low Level Execution and Feedback} describes how these object-centric targets are translated into robot trajectories and how execution results are fed back to the task planner. 
Algorithm \ref{alg} summarizes the overall workflow of our framework.

\begin{algorithm}
\caption{AdaptPNP}
\label{alg}
\begin{algorithmic}[1]
\REQUIRE Initial observation \(\mathcal{O}_I\), task instruction \(\mathcal{I}\), action primitives \(\mathbf{S}\)
\ENSURE Task succeeds
\STATE \(\mathit{success} \leftarrow \text{False},\ \mathcal{O} \leftarrow \mathcal{O}_I\)
\WHILE{\(\neg\,\mathit{success}\)}
  \STATE \textbf{1. Plan Skeleton Generation}  
  \STATE \(\mathcal{P} \leftarrow \text{TaskPlanner}(\mathcal{O}, \mathcal{I}, \mathbf{S})\)
  \STATE \textbf{2. Sub-goal Pose Generation}  
  \FOR{each action primitive \(s_i \in \mathcal{P}\)}
    \STATE Pose sampling in digital twin \(\mathcal{D}\)
    \STATE VLM selection of best sub-goal \(\{T^{\text{tgt}}_i\}\)
  \ENDFOR
  \STATE \textbf{3. Execution \& Feedback}  
  \STATE Map \(\{T^{\text{tgt}}_i\}\) to execution environment \(\mathcal{E}\) and execute
  \IF{all executions succeed}
    \STATE \(\mathit{success} \leftarrow \text{True}\)
  \ELSE
    \STATE Receive error \(e\) and new observation \(\mathcal{O}\)
    \STATE \(\mathcal{P} \leftarrow \text{Replan}(\mathcal{O}, \mathcal{I}, e, \mathcal{P})\)
  \ENDIF
\ENDWHILE
\RETURN \(\mathit{success}\)
\end{algorithmic}
\end{algorithm}

\subsection{Plan Skeleton Generation and Reflection}
\label{Sec: Plan Skeleton Generation and Reflection}

To address adaptive task planning in hybrid P\&NP manipulation scenarios, our framework employs two complementary VLM modules: a Task Planner for task plan generation and modification; a Reflector for reasoning execution result and feeding back to task planner.

\subsubsection{Task Planner}  
Given a visual observation $\mathcal{O}_{I}$ (typically a third-person top-down RGB image) and a textual task instruction $\mathcal{I}$, the VLM Planner generates an initial plan skeleton 
$\mathcal{P}_{I} = [s_1({\lambda_{1}} ), \dots, s_m({\lambda_{m}})]$. Prompts include natural-language definitions of all action primitives $\mathbf{S} = \{s_{1}, \ldots, s_{n}\}$ with their effects and preconditions, inspired by PDDL syntax~\cite{aeronautiques1998pddl}. We employ Chain-of-Thought reasoning~\cite{wei2022chain} to systematically infer object identities, spatial relations, and environment features from $\mathcal{O}_{I}$, integrating them with $\mathcal{I}$ and $\mathbf{S}$ to reduce hallucinations and enhance plan coherence.
Leveraging the VLM’s strong visual comprehension and zero-shot generalization, the planner can recognize diverse scene elements and propose plausible action sequences across novel objects and layouts. Notably, this initial plan relies exclusively on visual semantics, deferring physical feasibility to the subsequent reflection loop.

\subsubsection{VLM Reflector}
Since P\&NP tasks exhibit multi-modal solution spaces where semantically plausible plans may be physically infeasible, we introduce a reflection mechanism to iteratively refine the task plan based on execution outcomes. After the initial skeleton $\mathcal{P}_{I}$ is populated with continuous parameters and executed by downstream modules (detailed in~\Cref{Sec: Sub-goal Pose Generation,Sec: Low Level Execution and Feedback}), the VLM Reflector analyzes any execution failures. When a plan step fails, the reflector receives a textual error message $e$ from low-level module describing the failure mode (for example, downstream module may raise ``Unable to solve an IK solution'')  and an updated visual observation $\mathcal{O}_{\text{new}}$ capturing the post-execution state. 
The reflector then performs reasoning to identify the root cause and provides feedback to the VLM Planner, which generates a revised plan skeleton. This reflection loop continues until a physically executable plan is achieved.

\subsection{Sub-goal Pose Generation}
\label{Sec: Sub-goal Pose Generation}




Given a plan skeleton $\mathcal{P}$, we must map it to a continuous robot trajectory $\mathcal{T}$. 
To do so, we introduce a ``mental rehearsal'' stage in which a digital twin generates candidate 6D object target poses for each $s_i(\lambda_i) \in \mathcal{P}$. These sub-goal poses serve as a fine-grained intermediate representation between high-level task planning and low-level motion execution, capturing physically feasible object transformations that downstream controllers can follow directly.

We divide this process into two steps: first, \textbf{Pose Sampling} in the digital twin produces a small set of feasible candidate poses; second, \textbf{VLM Selection} uses visual prompts of these poses to pick the most semantically and physically appropriate target as sub-goal pose for execution.

\subsubsection{Pose Sampling}
Grounding VLM reasoning in 3D space is challenging, so we combine foundation models with digital-twin validation to sample candidate poses:
\begin{itemize}
  \item \textbf{Reasoning to Region.} A VLM (GPT-4o) identifies the target region and 
further refines it into a detailed and positional description derived from the plan primitive \(s_i(\lambda_i)\) (e.g., transforming ``table edge'' into 
``the nearest {table edge} to the \{object\}'').
  \item \textbf{2D Grounding.} Seed1.5-VL~\cite{guo2025seed1} localizes this description to a pixel coordinate \((x_i,y_i)\) in the observation image, which is then projected via the camera matrix to a 3D world point \((x_w,y_w,z_w)\).
  \item \textbf{Pose Generation.} Around \((x_w,y_w,z_w)\), we sample candidate poses \(\{T_j\}\) according to the action type’s allowed degrees of freedom ($e.g.$, \((x,y,\text{yaw})\) for pushes). Each pose is instantiated with setting object in the pose in digital twin and simulated until it settles.
  \item \textbf{Feasibility Filtering.} Infeasible samples ($e.g.$, objects toppled or fallen off the table) are discarded. Remaining poses are ranked by reachability, and the top four are rendered as visual prompts \(\mathcal{O}_s = \{\mathcal{O}_1,\dots,\mathcal{O}_4\}\).
\end{itemize}
\subsubsection{VLM Selection}
Given the rendered prompts \(\mathcal{O}_s\), along with the current skill \(s_i\) and next skill \(s_{i+1}\), the VLM selects the best candidate by outputting the index \(k\) of the chosen prompt \(\mathcal{O}_k\). The corresponding 6D pose \(T_k\) becomes the sub-goal \(T_{k}^{\text{tgt}}\) for current primitive low-level execution.

By decoupling semantic reasoning (VLM) from physical validation (digital twin), our method generates rich \(SE(3)\) spatial representations that generalize across objects, tasks and policies while guaranteeing physical feasibility.

\subsection{Low-Level Execution and Feedback}
\label{Sec: Low Level Execution and Feedback}

In this stage, we map the initial object poses \(\{T^{\text{init}}_1, \dots, T^{\text{init}}_k\}\) and target sub-goal poses \(\{T^{\text{tgt}}_1, \dots, T^{\text{tgt}}_k\}\) generated in the digital twin \(\mathcal{D}\) to robot actions in the execution environment \(\mathcal{E}\) (either real or simulated) and use the execution outcomes to refine the task plan. By using these fine-grained 6D poses poses, various low-level control policies, such as MPC or RL, can be applied directly.

However, discrepancies in dynamics and physical properties between \(\mathcal{D}\) and \(\mathcal{E}\) often cause actions that succeed in the digital twin to fail in execution environment. This gap is especially problematic when using non-prehensile manipulation skills, where the absence of stable grasping makes it difficult to correlate robot motions with object motions.
To mitigate this, we rely on the digital twin only for sampling object poses, not robot trajectories. The system communicates only object sub-goal poses \(\{T^{\text{tgt}}\}\) and action primitive $s_i(\lambda_i)$ to \(\mathcal{E}\), rather than explicit joint commands. This object-centric interface sidesteps the physics gap, since it abstracts away robot dynamics, and thus supports cross-embodiment transfer across different robots.

Assuming \(\mathcal{D}\) and \(\mathcal{E}\) are well aligned, we directly map each pose \(T^{\text{init}}\) and \(T^{\text{tgt}}\) to \(\mathcal{E}\). Given that the task planner has already reduced the action space to a specific primitive \(s_i\), we implement heuristic policies for action primitives to drive the object from 
$T^{\text{init}} = (x_i,y_i,z_i, \text{roll}_i,\text{pitch}_i,\text{yaw}_i)$ to $T^{\text{tgt}} = (x_t,y_t,z_t, \text{roll}_t,\text{pitch}_t,\text{yaw}_t)$.

\textbf{Push.} Constrained to \(SE(2)\), we align \((x_i,y_i)\) with \((x_t,y_t)\) by pushing the object along the direction vector \(\mathbf{v}=(x_t-x_i,\,y_t-y_i)\) pointing from the object to the target center. Yaw directional alignment uses Farthest Point Sampling (FPS) to sample contact points and pushes along their normals to rotate. Both phases run under a PID controller until the alignment error falls below a threshold.

\textbf{Rotate.} We identify the support edge of a 3D bounding box that is closest to the target and apply forces on the opposite edges’ normals to induce rotation about that edge, aligning the object’s orientation with the target's.

\textbf{Grasp, Moveto, and Release.} For \textit{Grasp}, we use Anygrasp~\cite{fang2023anygrasp} to compute a grasp pose and a motion planner~\cite{cheng2018rmp} to execute the grasp. \textit{Moveto} and \textit{release} can be simply done by motion planner and robot controller.

If the execution failed, robot controller can directly detect and report error types, such as IK failure, collision, or out-of-reach errors. As a low-level module, Anygrasp can also provide low-level feedback when no grasp pose found. Subsequently, the error message $e$ combined with the new observation $\mathcal{O}$ is passed to the VLM reflector to reason a likely cause of failure. This reasoning result together with $\mathcal O$ are ultimately passed to VLM planner to enhance replan.

\section{experiment}

In this section, we perform a series of experiments aimed at addressing the following questions.
\begin{itemize}
    \item Can our framework effectively address diverse manipulation tasks that intrinsically require a hybrid of prehensile and non-prehensile skills? (\Cref{exp: simulation})
    \item How much do our two key designs, 6D object pose intermediate representation and the closed-loop reflection mechanism, contribute to performance? (\Cref{exp: ablation})
    \item Does our approach maintain its effectiveness when applied to real-world scenarios? (\Cref{exp: real})
\end{itemize}

\begin{figure}[!ht]
\centering
\includegraphics[width=1.0\linewidth]{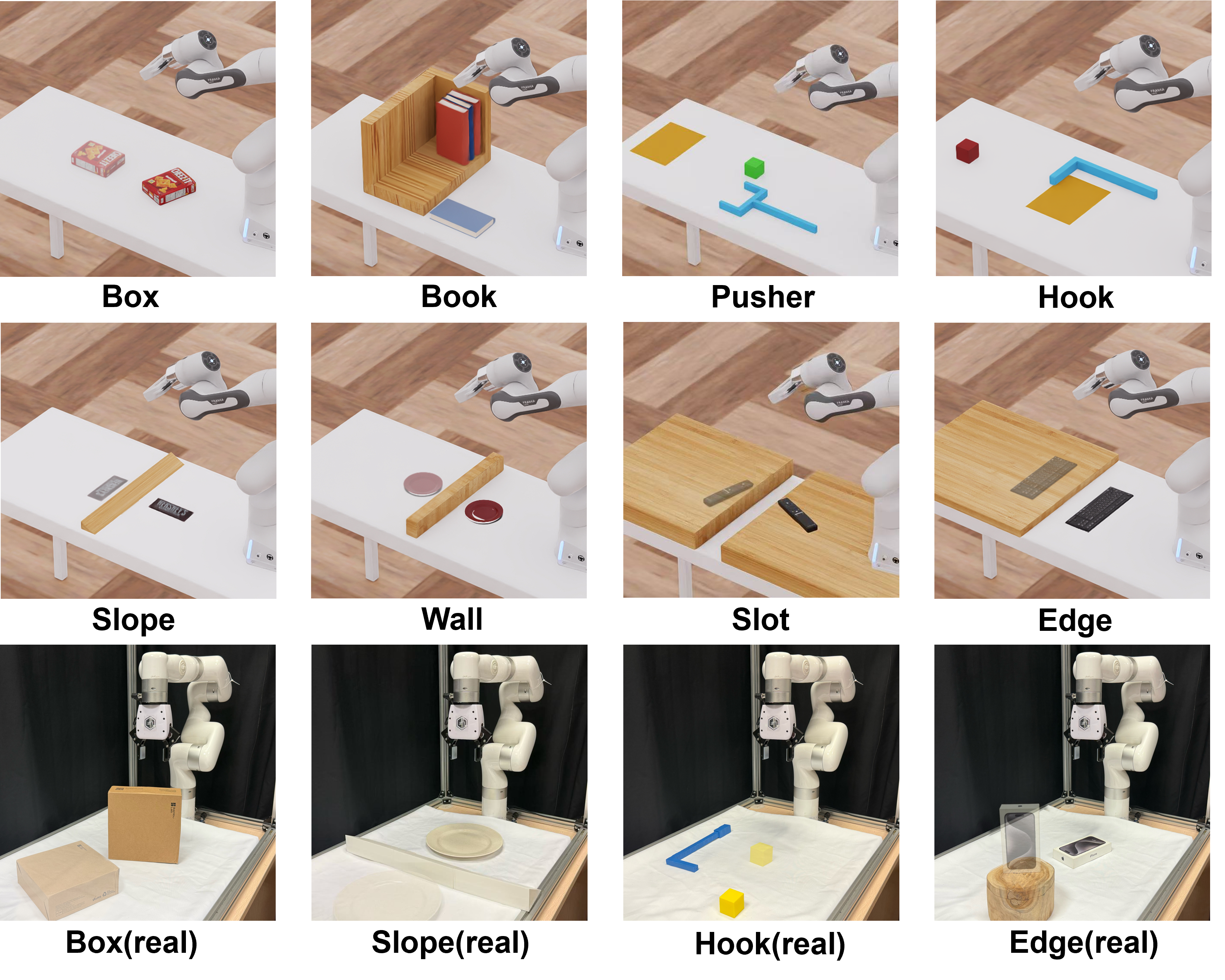}
\caption{\textbf{Task Setup.} We evaluate \ours\ on a spectrum of P\&NP hybrid manipulation scenarios, including eight simulated tasks (top two rows) and four real-world tasks (bottom row). In each scene, the final target pose is shown as a translucent object, and the target region is indicated by a yellow overlay ($e.g.$, Pusher, Hook). }
\label{fig:task}
\end{figure}
\subsection{Simulation Experiment}
\label{exp: simulation}



{
\setlength{\tabcolsep}{7.6mm}{
\begin{table*}[!ht]
    \centering
 \caption{\textbf{Task Description}}
    \begin{tabular}{lll}  
    \toprule

    Tasks & Type & Feature  \\
    \midrule
    Box   &  Alignment   & Object too wide, need to push or rotate to align.\\
    Book  & Alignment & Highly constrained environment that needs to rotate out the book.  \\
    Edge & Extrinsic Dexterity  & Object too thin, need to leverage the table edge to facilitate grasping. \\
    Wall   &  Extrinsic Dexterity  & Object too thin, need to leverage the table edge or wall to facilitate grasping. \\
    Slope    &  Extrinsic Dexterity  & Object too thin, need to leverage the table edge or slope to facilitate grasping.  \\
    Slot   &   Extrinsic Dexterity & Object too thin, need to leverage the table edge or slot to facilitate grasping. \\
    Tool Hook     &  Tool Using  & Object out of reach, need to use a hook to pull the object back to target. \\
    Tool Pusher     &  Tool Using  & Target out of reach, need to use a pusher to push the object to target. \\
    \bottomrule
    \end{tabular}
    \label{tab:tasks}
\end{table*}
}
}

\begin{table*}[!t]
    \centering
    \setlength\tabcolsep{6pt} 
    \renewcommand{\arraystretch}{1.2}
    \begin{threeparttable}
    \captionsetup{width=\linewidth}
    \caption{\textbf{Quantitative Results of Simulation Tasks} ($\uparrow$)}
    \label{tab:sim result}
    \begin{tabular}{@{}lcccccccc@{}} 
    \toprule
    & Box & Book & Edge & Wall & Slope & Slot & Tool Hook& Tool Pusher\\ 
    \midrule
    MPC      &  8/10 & 1/10  &  0/10 & 0/10  & 0/10  & 0/10   & 0/10  &0/10  \\
    RL       & 9/10  & 3/10  &  0/10 &  0/10 & 0/10  & 0/10  &0/10   & 0/10 \\
    MOKA     & 2/10  & 0/10  & 1/10  & 3/10  & 2/10  & 1/10  & 1/10 & 0/10 \\
    MolmoAct & 3/10  & 0/10  & 0/10  & 0/10  & 1/10 & 0/10  & 0/10 & 0/10 \\
    OpenVLA      & 1/10  & 0/10  & 0/10  & 0/10  & 0/10  & 0/10  & 0/10 & 0/10 \\
    \rowcolor{customblue} \textbf{Ours} 
             & \textbf{9/10} & \textbf{7/10} & \textbf{6/10} & \textbf{8/10} & \textbf{5/10} & \textbf{9/10} & \textbf{6/10} & \textbf{3/10} \\
    \bottomrule
    \end{tabular}
    \end{threeparttable}
    \vspace{-2mm}
\end{table*}



\subsubsection{Task Selection}
To comprehensively evaluate our framework, we select eight representative hybrid P\&NP manipulation tasks and implement them in IsaacSim~\cite{NVIDIA_Isaac_Sim} as a standardized benchmark. We categorize these tasks into three groups: (1) alignment, (2) extrinsic dexterity, and (3) tool using, based on the task feature. 
In most scenarios, the objects are purposely ungraspable initially ($e.g.$, too thin, flush with the surface, or outside the gripper’s aperture), intrinsically requiring the robot to flexibly adopt prehensile and non-prehensile skills to achieve the goal.
\Cref{tab:tasks} provides detailed description of each task, while~\Cref{fig:task} illustrates the visual setup of these manipulation scenarios.

\subsubsection{Experiment Setup}
We conduct ten independent trials for each task in IsaacSim~\cite{NVIDIA_Isaac_Sim}. For tool using manipulation tasks, success is defined as placing the target object into the designated region. For alignment and extrinsic dexterity tasks, each object is assigned a 6D target pose, and a trial is considered successful if the Euclidean distance between the object center and the target center is less than 3 cm, and the orientation error, measured as the geodesic distance between the current and target quaternions, is less than 10 degrees. To ensure robustness, we randomize object initial conditions in each trial by applying a uniform perturbation of $\pm 5$\,cm in position ($x$, $y$) and $\pm 30^\circ$ in orientation. For the box task, we additionally consider two initial placement states, standing and lying, with five trials each. The target pose is also randomized with positional and orientational noise. Across all tasks, a task is considered as successful if the criteria is met within three times of replans.

\subsubsection{Baseline}
We compare our method against classical control, reinforcement learning, hierarchical VLM-based planning, and end-to-end vision-language-action models
\begin{itemize}
\item \textbf{MPC:} Model Predictive Path Integral (MPPI)~\cite{williams2017information} in IsaacLab, which optimizes short-horizon trajectories in parallel.
\item \textbf{RL:} Proximal Policy Optimization (PPO)~\cite{schulman2017proximal}, trained in IsaacLab with the 6D target pose error as reward.
\item \textbf{MoKA}~\cite{liu2024moka}: a VLM-based method that decomposes instructions into subtasks and selects 2D keypoints as intermediate representations.

\item \textbf{MolmoAct}~\cite{lee2025molmoact}: generate waypoints and robot trajectories according to instruction in 2D image.

\item \textbf{OpenVLA}~\cite{kim2024openvla}: a 7B end-to-end vision language action model trained on 970k real-world demonstrations.

\end{itemize}
We use the same success criteria to evaluate the performance of our method and all baselines.

\subsubsection{Result Analysis}
As shown in~\Cref{tab:sim result}. Both \textbf{MPC} and \textbf{RL} perform relatively well on simple tasks such as Box Alignment, which require only a few steps to complete. 
Sample-based MPC benefits from its ability to perform short-horizon optimization with parallel environments as the predictive model, enabling it to rapidly converge to a feasible trajectory. Similarly, RL can efficiently discover rewarding action sequences in this low-dimensional, dense-reward setting, quickly receiving positive feedback and adjusting its policy to align the object with the target. However, their performance degrades significantly on long-horizon tasks, especially in environments with non-monotonic reward landscapes, where the optimal strategy requires temporarily reducing the reward signal, such as in the Wall task where the object needs to detour around an obstacle to reach the target. Such long-horizon tasks challenge both MPC and RL due to their preference for locally greedy solutions. In contrast, our pipeline adopt the task and motion planning formulation, decompositing a complex task into several subtasks, which reduces the search space and significantly eases the burden on the low-level policy to find a feasible solution.

\textbf{MoKa} uses a VLM to break down detailed task instructions into subtasks, reducing the action space and enabling long-horizon planning. We extend MoKa to include NP primitives, but it still depends solely on 2D points as intermediate representations. Similarly, \textbf{MolmoAct} use 2D trajectory as representation for low-level execution. Both of their limitation in 2D representation hamper execution of fine-grained motions, such as rotations, causing poor performance on tasks like Box and Book Alignment.

We also evaluate the end-to-end language-action model \textbf{OpenVLA} in a zero-shot setting. Across all tasks in our experiments, OpenVLA performs poorly, highlighting its limited generalization to novel tasks beyond its training distribution. This performance degradation is especially pronounced on our complex P\&NP hybrid manipulation tasks, underscoring the challenges these tasks pose for current end-to-end language-action models.

In contrast, our approach achieves the highest success rate across all tasks. We attribute this performance to three key innovations. First, our framework enables flexible composition of prehensile and non-prehensile primitives, essential for scenarios where objects cannot be directly grasped or require pre-manipulation to become accessible. 
Second, we employ 6D object poses as intermediate representations, capturing full spatial information that enables precise control over fine-grained actions like rotation and directional pushing. 
Third, our closed-loop reflection mechanism iteratively refines the plan skeleton based on execution feedback, proving crucial for long-horizon tasks with complex constraints. 
Together, these designs significantly reduce the search burden on low-level policies, enhance robustness in constrained environments, and deliver consistent superior performance.




\begin{figure*}[!htb]
\centering
\includegraphics[width=1.0\linewidth]{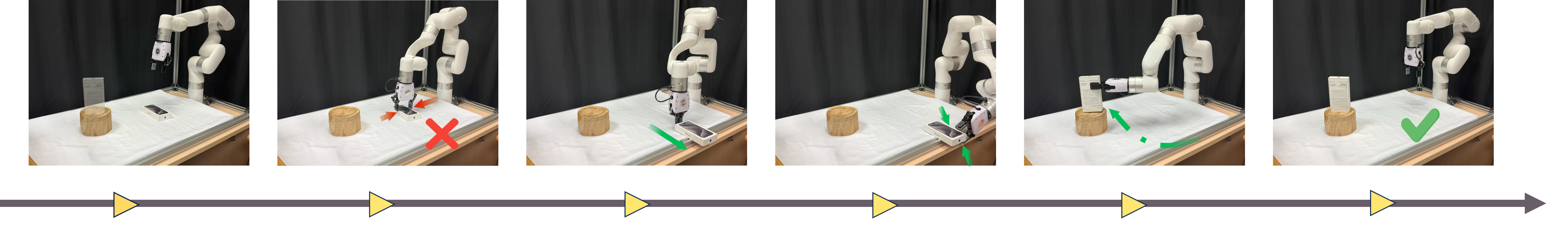}
\caption{\textbf{Real-World Task Process Visualization.} The goal is to place the box at the translucent target pose. Direct grasp fails because the box is slightly wider than the gripper. \ours\ replans by first pushing the box to the table edge and then grasping it from the side, successfully reaching the goal. }
\label{fig:real}
\end{figure*}

\subsection{Ablation Study}
\label{exp: ablation}
\begin{table}[!ht]
    \centering
    \setlength\tabcolsep{4pt} 
    \renewcommand{\arraystretch}{1.0} 
    \begin{threeparttable}
    \captionsetup{width=0.9\linewidth}
    \caption{\textbf{Ablation Study of Our Method} ($\uparrow$)}
    \begin{tabular}{@{}lccccccc@{}} 
    \toprule
    & Book & Wall& Slot & Tool Hook \\ 
    \midrule
    Ours(w/o pose)& 3/10& 2/10& 4/10& 1/10& \\
    Ours(w/o reflection) & 0/10& 1/10 &0/10 &0/10 \\
    \rowcolor{customblue} \textbf{Ours} & \textbf{7/10} & \textbf{8/10} & \textbf{9/10} &\textbf{6/10}  & \\ 
    \bottomrule
    \label{tab:ablation}
    \end{tabular}
    \end{threeparttable}
    \vspace{-2mm}
\end{table}

We evaluate the contributions of our two key designs: (1) using 6D object poses as the intermediate representation, and (2) the closed-loop task plan reflection mechanism.
\subsubsection{Effect of 6D Pose Representation}
To isolate the impact of pose-based sub-goals, we replace our 6D poses with 2D keypoints plus depth (as in prior works~\cite{liu2024moka, huang2024rekep, yuan2024robopoint}), and simplify rotation to a fixed 90° turn toward the target. Results in Table~\ref{tab:parts_performance} (row 1) show large performance drops, especially on extrinsic dexterity tasks (Wall, Slot). Without physics-informed 6D grounding, the VLM cannot accurately predict stable sub-goal configurations, leading to objects slipping or misoriented placements. Moreover, 2D points fail to capture full translation and rotation nuances, limiting fine-grained control. Finally, visualizing full 6D pose candidates provides richer context for the VLM, particularly in tool using tasks where both hook reachability and spatial alignment matter.

\subsubsection{Effect of Reflection Mechanism}
We next disable the iterative feedback loop, forcing a one-shot plan without reflection. As shown in Table~\ref{tab:ablation} (row 2), almost all tasks fail across ten trials.
Our tasks, featuring blocked objects (Book), thin objects (Wall, Slot), and out-of-reach goals (Tool Hook), demand adaptive sequencing of NP actions to render grasping possible. However, only RGB observation cannot provide sufficient information for VLM to understand the non-trival features of the scene. Without execution feedback, the VLM planner lacks physical feasibility checks and resorts to direct grasp-and-move strategies that are infeasible. In contrast, with reflection enabled, our method iteratively refines semantically plausible plans into physically executable ones, dramatically improving success rates and demonstrating the critical role of closed-loop feedback.

\subsection{Real World Experiment}
\label{exp: real}

To validate our pipeline’s consistency and effectiveness beyond simulation, we evaluate four tasks, Box, Edge, Slope, and Tool Hook, in a real-world setup (\Cref{fig:task}). We use a UFactory xArm6 manipulator and an Intel RealSense D435 camera for RGB input. The digital twin is constructed in APP Reality Composer, and FoundationPose~\cite{wen2024foundationpose} provides real-time 6D pose estimates to align objects between the real scene and the twin. 
\Cref{fig:real} visualizes the process of Edge task, illustrating how our method adaptively refines the plan based on environmental features.

At runtime, we capture an RGB image to generate the initial plan skeleton. For each sub-goal, we extract a 2D target point $(x_i,y_i)$ in the image plane as described in ~\Cref{Sec: Sub-goal Pose Generation}, then back-project it to a 3D point $(x_w,y_w,z_w)$ using camera intrinsics $K_I$ and extrinsics $K_E$. This 3D point anchors the corresponding digital twin pose. After sampling and selecting the sub-goal pose in the twin, we execute the associated primitive in the real world using the heuristic policies from~\Cref{Sec: Low Level Execution and Feedback}. We compare our approach to two zero-shot baselines: MoKa and OpenVLA.

\begin{table}[!htb]
    \centering
    \setlength\tabcolsep{4pt} 
    \renewcommand{\arraystretch}{1.2} 
    \begin{threeparttable}
    \captionsetup{width=\linewidth}
    \caption{\textbf{Real-world Experiment Result} ($\uparrow$)}
    \label{tab:parts_performance}
    \begin{tabular}{@{}lcccccc@{}} 
    \toprule
    & Box & Edge& Slope & Tool Hook \\ 
    \midrule
    OpenVLA & 0/10& 0/10& 0/10& 0/10 \\
    MOKA& 2/10& 1/10& 1/10& 1/10  \\
    \rowcolor{customblue} \textbf{Ours} & \textbf{8/10} & \textbf{4/10} & \textbf{3/10} & \textbf{5/10}\\ 
    \bottomrule
    \label{tab:realworld}
    \end{tabular}
    \end{threeparttable}
    \vspace{-2mm}
\end{table}

\Cref{tab:realworld} reports success rates. Both OpenVLA and MoKa perform poorly under zero-shot conditions due to the challenging task geometries, mirroring their simulation failures. In contrast, our pipeline achieves success rates comparable to simulation, significantly outperforming both baselines and demonstrating robust sim-to-real transfer. A key advantage of our approach is using 6D sub-goal poses as the interface between the digital twin and execution environment: transferring object poses rather than robot commands bypasses the sim-to-real gap and yields consistent real-world performance.



\section{conclusion}

We presented AdaptPNP, a unified framework for adaptive task-and-motion planning that integrates prehensile and non-prehensile manipulation within a single system. A VLM-based planner generates high-level plan skeletons, which are grounded through 6D target pose sampling in a digital twin. A closed-loop reflection mechanism further refines plans based on execution feedback, enabling adaptation to object properties and environmental constraints.

Extensive experiments on representative hybrid P\&NP tasks show that AdaptPNP outperforms competitive baselines in both success rate and robustness. By bridging semantic planning and low-level execution through 6D pose representations and iterative refinement, AdaptPNP advances adaptive manipulation in diverse and unstructured environments.

\small
\bibliographystyle{IEEEtran}
\bibliography{references}




\end{document}